\documentclass[letterpaper, 10 pt, conference]{ieeeconf}  

\IEEEoverridecommandlockouts                              

                                       \usepackage{graphics} 
\graphicspath{ {images} }
\usepackage{epsfig} 
\usepackage{mathptmx} 
\usepackage{times} 
\usepackage{amsmath} 
\usepackage{amssymb}  
\usepackage[noadjust]{cite}
\usepackage[hidelinks]{hyperref}
\usepackage{threeparttable}
\usepackage{multirow}
\usepackage{hhline}
\usepackage{todonotes}
\usepackage{epstopdf}
\newcommand\Tstrut{\rule{0pt}{2.0ex}}         

\title{\LARGE \bf
Deep Fruit Detection in Orchards
}

\author{Suchet Bargoti and James Underwood$^{1}$
\thanks{$^{1}$The authors are with the Australian Centre for Field Robotics, The University of Sydney, 2006, Australia. {\tt\small s.bargoti, j.underwood   @acfr.usyd.edu.au}}
}

\begin{document}

\maketitle

\begin{abstract}
An accurate and reliable image based fruit detection system is critical for supporting higher level agriculture tasks such as yield mapping and robotic harvesting. This paper presents the use of a state-of-the-art object detection framework, Faster R-CNN, in the context of fruit detection in orchards, including mangoes, almonds and apples. Ablation studies are presented to better understand the practical deployment of the detection network, including how much training data is required to capture variability in the dataset. Data augmentation techniques are shown to yield significant performance gains, resulting in a greater than two-fold reduction in the number of training images required. In contrast, transferring knowledge between orchards contributed to negligible performance gain over initialising the Deep Convolutional Neural Network directly from ImageNet features. Finally, to operate over orchard data containing between 100-1000 fruit per image, a tiling approach is introduced for the Faster R-CNN framework. The study has resulted in the best yet detection performance for these orchards relative to previous works, with an F1-score of $>0.9$ achieved for apples and mangoes. 
\end{abstract}

\section{INTRODUCTION}
\label{sec:intro}
Vision based fruit detection is a critical component for in-field automation in agriculture. With accurate knowledge of individual fruit locations in the field, it is possible to perform yield estimation and mapping, which is important for growers as it facilitates efficient utilisation of resources and improves returns per unit area and time. Precise localisation of the fruit is also a necessary component of an automated robotic harvesting system, which can help mitigate one of the most labour intensive tasks in an orchard \cite{Kapach:2012}.  

Typically, prior work utilises hand engineered features to encode visual attributes that discriminate fruit from non-fruit regions \cite{Payne2014,Wang2013,Nuske2014}. Although these approaches are well suited for the dataset they are designed for, feature encoding is generally unique to a specific fruit and the conditions under which the data were captured. More recently, advances in the computer vision community have translated to agrovision (computer vision in agriculture), achieving state-of-the-art results with the use of Deep Neural Networks (DNNs) for object detection and semantic image segmentation \cite{Sa2016,Bargoti2016JFR}. These networks avoid the need for hand-engineered features by automatically learning feature representations that discriminately capture the data distribution. 

\begin{figure}[!htb]
	\centering
	\includegraphics[width=0.48\textwidth,clip=true,trim={0 15 0 10}]{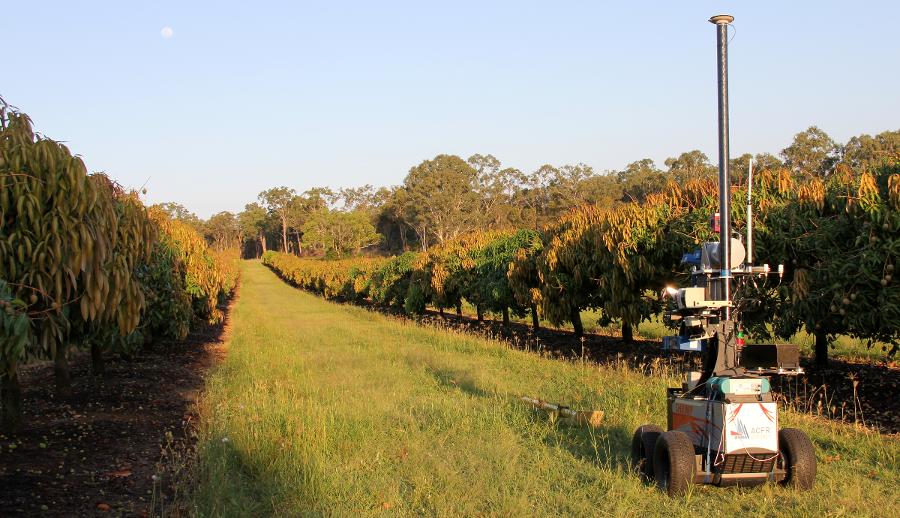}
	\caption{Research ground vehicle Shrimp, developed at the Australian Centre for Field Robotics, The University of Sydney, traversing between rows at a mango orchard, capturing tree image data.}
	\label{fig:shrimp-orchard}
\end{figure}

Outdoor orchard image data (being collected by a ground vehicle in Fig. \ref{fig:shrimp-orchard}) present additional challenges for fruit detection. For efficient large scale operation, sensor field of view needs to span entire trees, with high resolution imagery required for the relatively small fruit size. For example, almond tree image data used in this paper contains a large pixel and fruit count with $18$ megapixels (MP) images containing up to $1500$ fruit each. Additionally, as the data is captured in outdoor scenes, there is significant intra-class (within class) variations due to variability in: illumination conditions, distance to fruit, fruit clustering, camera view-point etc. These aspects result in a challenging data labelling process for supervised learning, and high resolution imagery imposes hardware/algorithm constraints.  

This paper address the specific constraints imposed by fruit detection in large scale orchard data, using a state-of-the-art deep learning detector, Faster R-CNN. The paper provides implementation details, rationale for design decisions, and ablation studies with experimentation spanning three significantly different orchard types, including apples, mangoes and almonds. The primary contributions are:
\begin{itemize}
\item Deploying a state-of-the-art object detection architecture, Faster R-CNN, in the context of fruit detection on outdoor orchard images. 
\item Empirical analysis on training data requirements to help minimise labelling efforts, through data augmentation proposals and transfer learning between orchards.
\item Proposing image modification strategies to perform detection on high resolution data containing more than $1000$ objects each. 
\item Releasing datasets\footnote{Accessible from \url{http://data.acfr.usyd.edu.au/ag/treecrops/2016-multifruit/}} used in this work and authors' previous publications \cite{Bargoti2016ICRA,Bargoti2016JFR,Hung2015}, alongside an object labelling annotation toolbox, designed for rapid fruit labelling \cite{Bargoti2016Labeller}. 
\end{itemize}

The remainder of the paper is organised as follows. Section \ref{sec:related-work} presents the related work for object detection in computer vision with a focus towards agrovision. Section \ref{sec:method} describes the detection approach, which uses the Faster R-CNN framework. Section \ref{sec:setup} details our experimental setup with the ablation studies presented in Section \ref{sec:results}. We discuss our findings and lessons learned in Section \ref{sec:discussion}, concluding in Section \ref{sec:conclusion} with future directions.  

\section{RELATED WORK}
\label{sec:related-work}

Fruit detection has been explored by many researchers in agrovision, across a variety of orchard types for the purposes of autonomous harvesting or yield mapping/estimation \cite{Bargoti2016JFR,Sa2016,Nuske2014,Kapach:2012}. Detection is typically performed by transforming image regions into discriminative features spaces and using trained classifiers to associate them to either fruit or background objects such as foliage, branches and ground. Semantic image segmentation performs this densely, resulting in a pixel-wise classification over the image. Post-processing techniques can then be applied to differentiate individual whole-objects of interest as groups of adjacent pixels. On the other hand, the detection search space can be reduced using low-level image analysis to identify regions of interests (RoIs) in the image (e.g. possible fruit regions), followed by high-level feature extraction and classification. 

Analysis of local colours and textures has been used for pixel-wise mango classification, followed by blob extraction to identify individual mangoes \cite{Payne2014}. Avoiding the use of hand-engineered features, Convolutional Neural Networks (CNNs) have been used for pixel-wise apple classification \cite{Bargoti2016JFR}, followed by Watershed Segmentation (WS) to identify individual apples. The radial symmetries in the specular reflection in berries have been shown to be useful for RoI extraction \cite{Nuske2014}, where a KD-forest was used for berry classification. To detect citrus fruit, Circular Hough Transforms (CHT) have been used to extract key-points, which were then classified using a Support Vector Machine (SVM) \cite{Sengupta2014}.

More recently, Region based Convolutional Neural Networks (R-CNN) \cite{Girshick2016}, which combine the RoI approach with CNNs, have produced state-of-the-art detection results on PASCAL-VOC detection dataset \cite{Everingham2010}. RoIs are initially proposed using Selective Search \cite{Uijlings2013}, which finds interesting regions merging superpixels. CNNs are used to classify the regions and directly regress a bounding box location for an object contained within. In subsequent work, the authors proposed the Faster R-CNN model \cite{Ren2015}, which merges region proposals and object classification and localisation into one unified deep object detection network. The end-to-end network yielded further improvements in detection results while significantly reducing the training and prediction times. 

In this paper we explore the necessary adaptations to the Faster R-CNN framework for fruit detection in orchard image data. Large scale orchard data is typically characterised by whole tree images containing thousands of fruit, with large variations in fruit sizes on the image data. Such data cannot be directly imported into the network due to hardware constraints and labelling images with a high object count can be a difficult task. We address the above issues and provide practical insights into data requirements, strategies for reducing training data and discuss knowledge transfer between orchards to aid fruit classification. This work is a natural progression of the authors' line of prior work in orchard imaging \cite{Bargoti2016JFR,Bargoti2016ICRA,Hung2013,Hung2015}, which has focused on achieving state-of-the-art results with fruit detection for orchard-scale yield mapping and estimation. 

Parallel to our work, \cite{Sa2016} recently demonstrated the use of Faster R-CNN for sweet pepper and rockmelon detection in a greenhouse and showed the versatility of the detector amongst 5 other fruit types with images obtained from Google Image search. Our orchard imagery differs substantially from that study, which warrants dedicated investigation. Greenhouses afford images taken relatively close to the fruit, and similarly the Google Images are taken with hand-held cameras. In both cases this leads to imagery with a relatively high pixel count per fruit and a low fruit-count per image when compared against outdoor orchard imagery (e.g. fruit in this paper have an average pixel width of $32\pm9$ pixels, compared to $112\pm29$ pixels for greenhouse fruit in \cite{Sa2016}). Furthermore, the larger trees and outdoor environment in orchards leads to greater illumination variability, despite our efforts to minimise this with strobe lights. We present additional details and guidance regarding how to structure Faster R-CNN for the particular requirements of orchard scale image data. Ongoing work on the Faster R-CNN framework has also been introduced recently, which advocates the use of Fully Convolutional Networks, resulting in further advances in accuracy and speed \cite{Zhang2016,Dai2016}. However, in this paper we focus on the original Faster R-CNN network \cite{Ren2015}, which is within the same family of deep learning based detectors, because of its easy-to-use open-source implementation.

\section{OBJECT DETECTION}
\label{sec:method}

\begin{figure*}[!ht]
	\centering
	\includegraphics[width=0.99\textwidth,clip,trim=0 10 0 18]{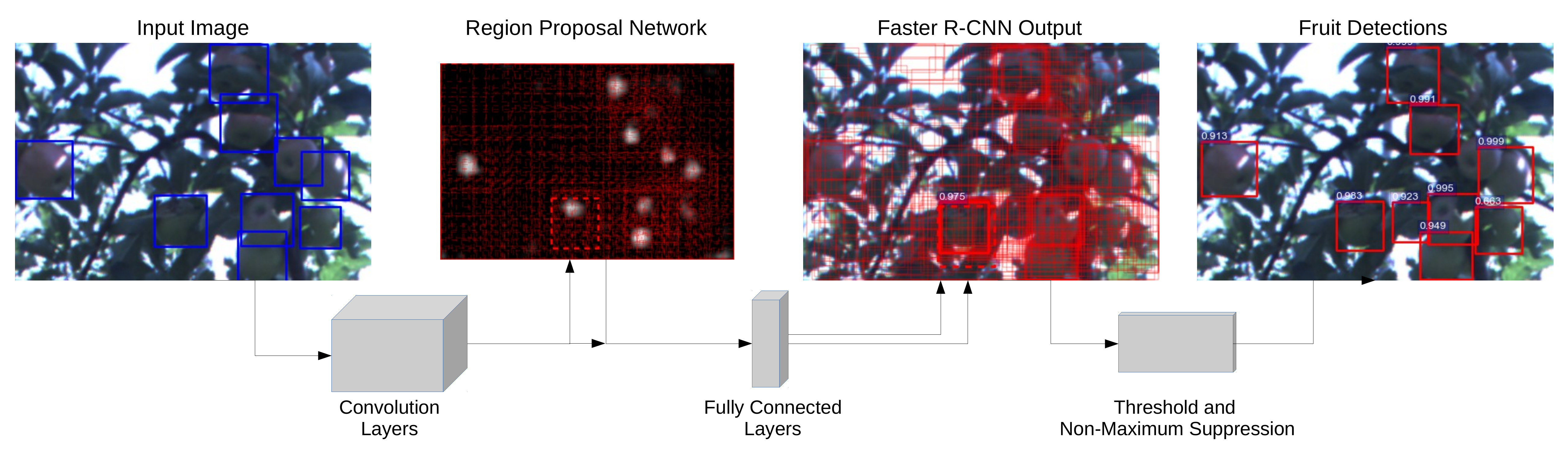}
	\caption{The Faster R-CNN Network. A 3-channel input image is propagated through a set of convolutional layers, from which Region of Interest boxes are proposed (dashed red boxes, with one high object probability box highlighted as an example). Each box is propagated through fully connected layers, which return their class probability and regresses a finer bounding box around individual objects (solid red boxes). Ground truth from the input image (in blue) is used in the RPN and the R-CNN layers during training. During testing, a class specific detection threshold is applied to the output, followed by Non-Maximum Suppression to remove overlapping results.}
	\label{fig:faster-rcnn}
\end{figure*}

This section presents the Faster R-CNN framework for fruit detection in orchards and introduces details about transfer learning and data augmentation techniques, which are used within the ablation studies conducted in Section \ref{sec:results}.

\subsection{Faster R-CNN}

The Faster R-CNN object detection system \cite{Ren2015} is composed of two modules: 1) a Region Proposal Network (RPN), used for detection of RoIs in the image, followed by 2) a classification module, which classifies the individual regions and regresses a bounding box around the object. 

During training, the input to the network is a 3-channel colour image (BGR) of arbitrary size (within constraints of GPU memory), along with annotated bounding boxes around each fruit. The image data is propagated through a number of convolutional layers depending on the choice of the CNN. In this paper we experiment with the ZF network, which contains 5 convolutional layers, and the deeper VGG16 net, which contains 13 convolutional layers (as done in \cite{Ren2015}). The output from the convolution layers is a high dimensional feature map, sub-sampled by a factor of $16$ due to the strides in the pooling layers. Local regions in the feature map are forward propagated into two sibling fully connected layers, a box-regression layer and a box-classification layer. This is the RPN layer, and the fixed number of class agnostic detections are the object proposals. Using attention mechanisms, individual proposals are propagated through subsequent fully connected layers (the R-CNN component), ending once again with two sibling layers with a finer region classification output and associated object bounding box. Training is done end-to-end using Stochastic Gradient Descent (SGD), allowing for the convolutional layers to be shared between the RPN and the R-CNN components. 

During testing, the network returns $N_p = 300$ bounding box detections per image (as in \cite{Ren2015}) with class probabilities. A probability threshold is applied, followed by Non-Maximum Suppression (NMS) to handle overlapping detections. The Faster R-CNN network is illustrated in Fig. \ref{fig:faster-rcnn}, showing intermediate outputs from a sample image from an apple orchard. The reader is referred to the original work for a detailed overview of the network architecture and other implementation details \cite{Ren2015}. 

\subsection{Transfer Learning}
It has become standard in computer vision to train a CNN using a large base network and then transfer (i.e. fine-tune) the learned features to a new target task, which typically has fewer labelled examples. The ImageNet dataset is often used as a base, containing 1000 object categories and 1.2 million images. Using ImageNet pre-trained CNN features, state-of-the-art results have been obtained on a variety of image processing tasks from image classification to image captioning \cite{Huh2016,Yosinski2014}.  

By default, Faster R-CNN advocates the initialisation of the detection network with weights learned from ImageNet \cite{Ren2015}. However, while transferring weights to a target task, performance can degrade if the target classes are drastically different to the base classes. This is because the deeper layers in a CNN network learn features that are more specific to the task at hand \cite{Yosinski2014}. This begs the question: what is the best way to initialise a network for the task of fruit detection in an orchard, where the data captured by a ground vehicle is typically very different from ImageNet? Should the networks be initialised by ImageNet features or would it more suitable to transfer knowledge from features fine-tuned over another orchard dataset?

\subsection{Data Augmentation}
Data augmentation is a common way to expand the variability of the training data by artificially enlarging the dataset using label-preserving transformations. The process increases the networks capability to generalise and reduces overfitting. Typical augmentation techniques used in the computer vision community include left-right flipping, image re-scaling, and changes to image colour. There are numerous approaches for colour augmentation, including colour/intensity jittering in a range of colour spaces such as RGB and HSV. This paper adapts the PCA augmentation technique presented in AlexNet \cite{Krizhevsky2012}, where the colour perturbations are along the natural variations in the dataset, denoted by the principal components of the pixel colours.

Augmentations can be implemented by either expanding the dataset with copies of the augmented versions, or by randomly augmenting the data during each training epoch. Employing the latter is preferable as it avoids pre-computing the wide range of random augmentations. 

\section{EXPERIMENTAL SETUP}
\label{sec:setup}


\begin{table*}[!t]
\footnotesize
\caption{Dataset configuration.}
\label{tab:dataset-info}
\begin{center}
\begin{threeparttable}{
\begin{tabular}{cccccccc}
\hline
Fruit & Sensor & Raw Img Size & Sub-Img Size & Fruit Width (px) & \#Fruit/Img & \#Train & \#Val/Test \Tstrut \\
\hline
Apple$^1$ & UGV + PointGrey LadyBug & $1616\times1232$ & $202\times308$ & $37\pm7$ & $4.5\pm2.9$ & $729$ & $112/112$ \Tstrut\\
Mango & UGV + Prosilica GT3300c & $3296\times2472$ & $500\times500$ & $34\pm11$ & $5.0\pm3.8$ & $1154$ & $270/270$\\
Almond & Handheld Canon EOS60D & $3456\times5184$ & $300\times300$ & $26\pm6$ & $7.4\pm5.6$ & $385$ & $100/100$\\ 
\hline
\end{tabular}
\begin{tablenotes}
\scriptsize
\item [1] Dataset previously used in \cite{Bargoti2016ICRA,Bargoti2016JFR,Hung2015}.
\end{tablenotes}}
\end{threeparttable}
\end{center}
\end{table*}

The orchard data evaluated in this paper consists of three fruit varieties: apples, almonds and mangoes, captured during daylight hours at orchards in Victoria and Queensland, Australia. The apple and mango data were captured with sensors on-board a general purpose research ground vehicle, built at the Australian Centre for Field Robotics (see Fig. \ref{fig:shrimp-orchard}). The vehicle traversed across different rows of the orchards collecting tree image data. The apple trees were trellised, enabling the ground vehicle to be in close proximity to the fruit. The longer distance between the mangoes and ground vehicle (illustrated in Fig. \ref{fig:shrimp-orchard}) was compensated for with a higher resolution sensor. Additionally, at the mango orchard, external strobe lighting was used with a small exposure time ($\sim70~\mu s$) to reduce variable illumination artefacts. Almond trees on the other hand have larger canopies and can host between $1000-10000$ almonds, which are of a smaller size than apples and mangoes. High resolution imagery was therefore required to obtain a good representation of the fruit, and was achieved with the use of a hand-held DSLR camera. Images captured in each orchard spanned entire trees, driven by the primary experimental objective of efficient yield estimation and mapping. The fruit detection work presented in this paper is therefore a critical component of the overarching project objective.  

The tree image data varied from $2-17$ MP, with each image containing around $100$ fruit for apples and mangoes, and over $1000$ fruit with almonds. However, hardware constraints limit the use of large images, with a $0.25$ MP image requiring $\sim2.5$ Gb of GPU memory with the VGG16 network. Additionally, for ground truth data collection, we found labelling a large number of small objects in large images to be a perceptually difficult task. We mitigate these problems by randomly sampling smaller sub-image patches from the pool of larger images acquired over the farm. This leaves us with smaller images (with a similar size as the PASCAL-VOC dataset used with Faster R-CNN) with low fruit counts, while covering the data variability across the farm. The data configurations for the different fruits is summarised in Table \ref{tab:dataset-info}. 

The ground truth fruit annotations for almonds and mangoes were collected using rectangular annotations, while circular annotations were more suitable for apples. However, Faster R-CNN operates on bounding box prediction, therefore the circular annotations were initially converted to rectangular ones of equal width and height. In practice, it was easier to label apples and mangoes than almonds, due to the size and contrast of the fruit compared to the surrounding foliage and the complexity of the canopy. To help differentiate the fruit from the background in shadowed regions of the image, the annotation software \cite{Bargoti2016Labeller}, provides sliders for contrast and brightness adjustments. 

Finally, the labelled dataset for each fruit was split into training, validation and testing splits (see Table \ref{tab:dataset-info}). The split was done such that each set contained data captured from a different part of the orchard block, in order to minimise biased results. Images in the training set that did not containing any fruit were discarded. 

\section{FRUIT DETECTION RESULTS}
\label{sec:results}
This section presents ablation studies for the detection network, assessing fruit detection performance with respect to the number of training images, transfer learning between orchards and data augmentation techniques. These studies were performed using the shallower ZF network as it is faster to train, however, the performance evaluation against the deeper VGG16 network is presented as well. Finally, a simple technique for deploying the learned networks over the large raw images (denoted as Tiled Faster R-CNN) is proposed. Although the Faster R-CNN framework is capable of multi-class detection, a binary problem is considered for orchard data, with a new model trained for each fruit type. Restricting the number of classes can generally lead to better classification accuracy \cite{Yosinski2014} and is acceptable in orchard applications as orchard blocks are typically homogeneous: one fruit per block without mixing \footnote{For pollination, different varieties may be interspersed, but the appearance variation is often less than for different fruit types.}. 

The ZF and VGG16 network have a sub-sampling factor of $16$ at the final convolution layer therefore the minimal possible object size is $16$ pixels. To ensure this, all training sub-images were scaled to have a shorter side of $500$ pixels, which meant enlarging the apple and almond sub-images. The sub-image dimensions specified in the previous section were chosen to allow for large enough fruit representations, post re-scaling. All networks were initialised with the ImageNet filters (unless otherwise stated) and trained until convergence in detection performance over the validation set. This was roughly $5000$ iterations for apples and almonds and $40000$ iterations for mangoes. Further exploration of initialisations and learning rates for the mango dataset could enable quicker training. All other network and learning hyper-parameters were fixed to the configuration used in \cite{Ren2015} for the PASCAL-VOC detection challenge. 

Detection performance is reported using the average-precision response for the fruits, the area under the precision recall curve. As in \cite{Bargoti2016JFR}, the final results are reported using F1-score, where the class threshold is evaluated over the held out validation set. The NMS threshold parameter was also optimised over the validation set and ranged between $0.2$ to $0.4$ for the different fruits, however, we found that the results were not sensitive in this range. A fruit detection was considered to be a true positive if the predicted and the ground truth bounding box had an Intersection over Union (IoU) greater than $0.2$. This equates to a $58\%$ overlap along each axis of the object, which was considered sufficient for a fruit mapping application. For example, with higher thresholds (such as $0.5$ used on PASCAL-VOC challenges), small errors in detections of smaller fruit caused them to be registered as false positives. A one-to-one matching was enforced during evaluation, penalising single detections over fruit clusters and multiple detections over a single fruit. 

\subsection{Number of Training Images}
\begin{figure}[!htb]
	\centering
		\includegraphics[width=0.48\textwidth,clip,trim=0 7 0 5]{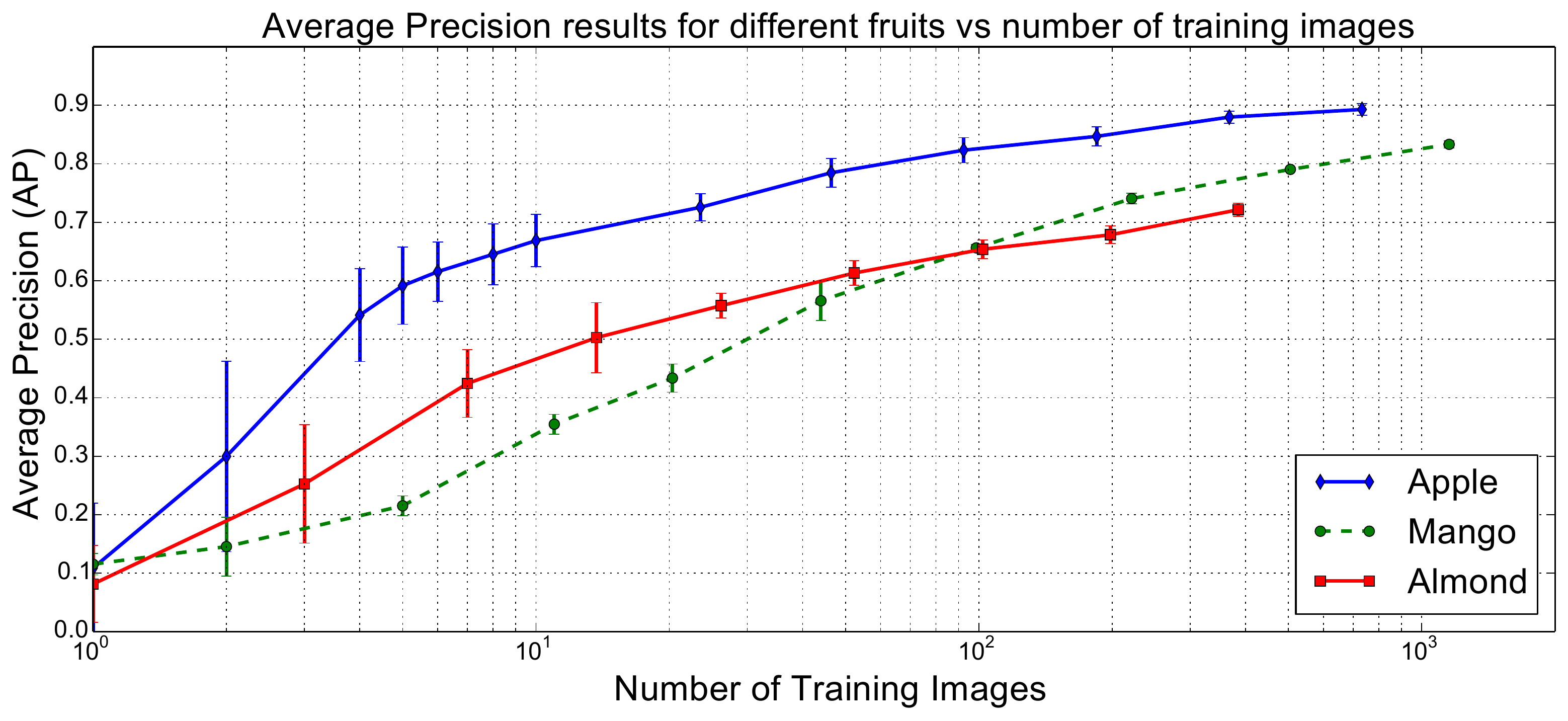}
		\caption{Detection performance for apples, mangoes and almonds as average precision vs number of training images.}
	\label{fig:num-training-images}
\end{figure}

For the three fruits, the number of training images in the learning phase were varied and the detection performance evaluated over the held out test set. The process was repeated 10 times to account for variance in the training data, where each time N random images were sampled from the training set without replacement. Fig. \ref{fig:num-training-images} shows the detection results for the three fruits as a function of number of training images. 

Detection performance rises quickly with a small number of training images, reaching $0.6$ for apples with just $5$ images. As the number of training images reaches the amount of available labelled data, performance is close to convergence for apples, only increasing by $0.01$ in the final x2 increase in the number of training images. The almond and mango data have not reached convergence yet, with both datasets yielding an improvement of more than $0.04$ AP in the last x2 increase in training data. 

\subsection{Transfer Learning}
\begin{figure}[!ht]
	\centering
		\includegraphics[width=0.48\textwidth,clip,trim=0 7 0 5]{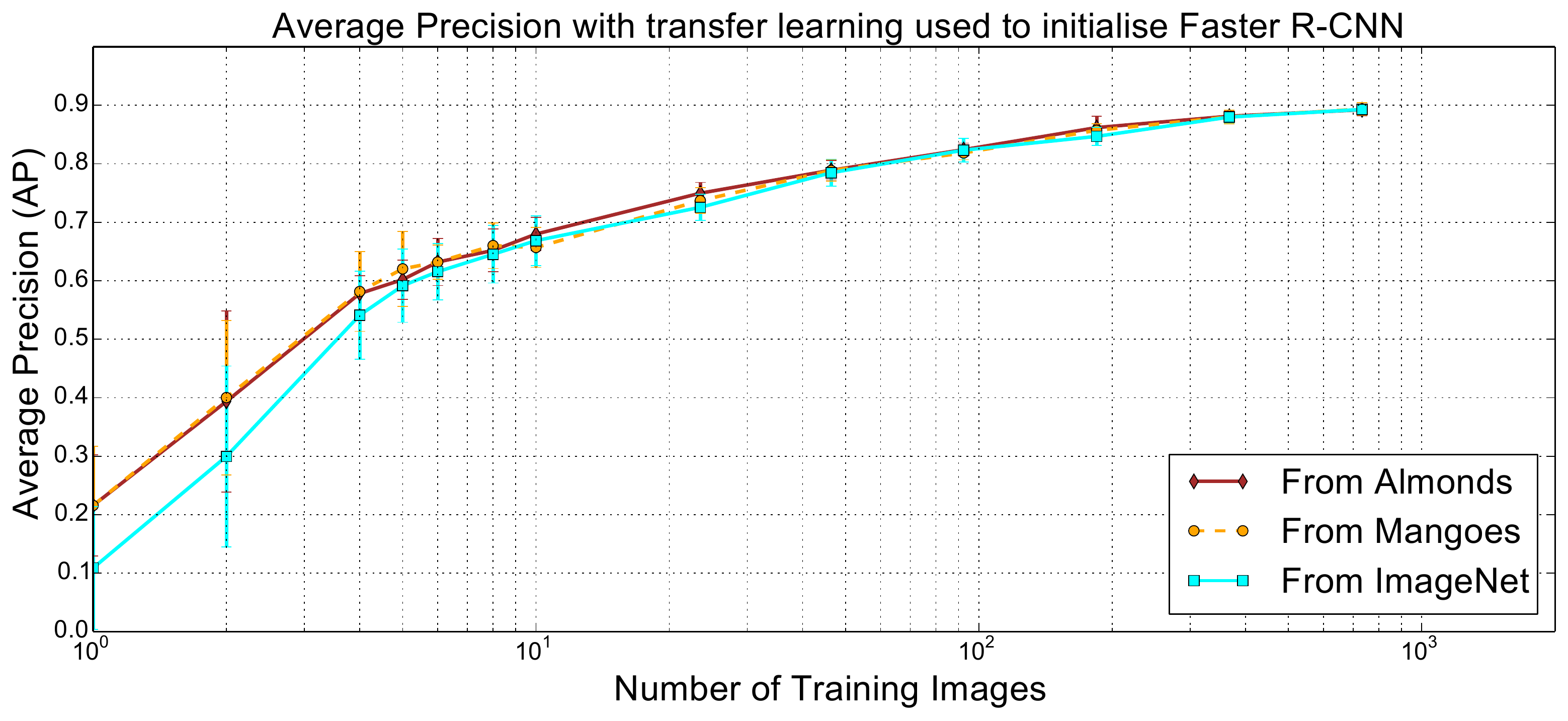}
		\caption{Apple detection performance with different transfer learning procedures. The default setting initialises the network weights during training with the ImageNet features. This is tested against networks initialised from features learned with almond and mango detection networks. }
	\label{fig:transfer-learning}
\end{figure}

To examine the utility of different types of transfer learning, the apple detection network was initialised using pre-trained models from mangoes and almonds (from the previous section). The detection performance vs number of training images is compared against a network initialised directly from ImageNet. The results (Fig. \ref{fig:transfer-learning}) show initial benefits with transfer learning from other orchards, which diminish quickly as the number of training images increase, with a difference of $0.01$ with just 5 training images. 

\subsection{Data Augmentation}
\begin{figure*}[!ht]
	\centering
		\includegraphics[width=0.98\textwidth,clip,trim=0 7 0 5]{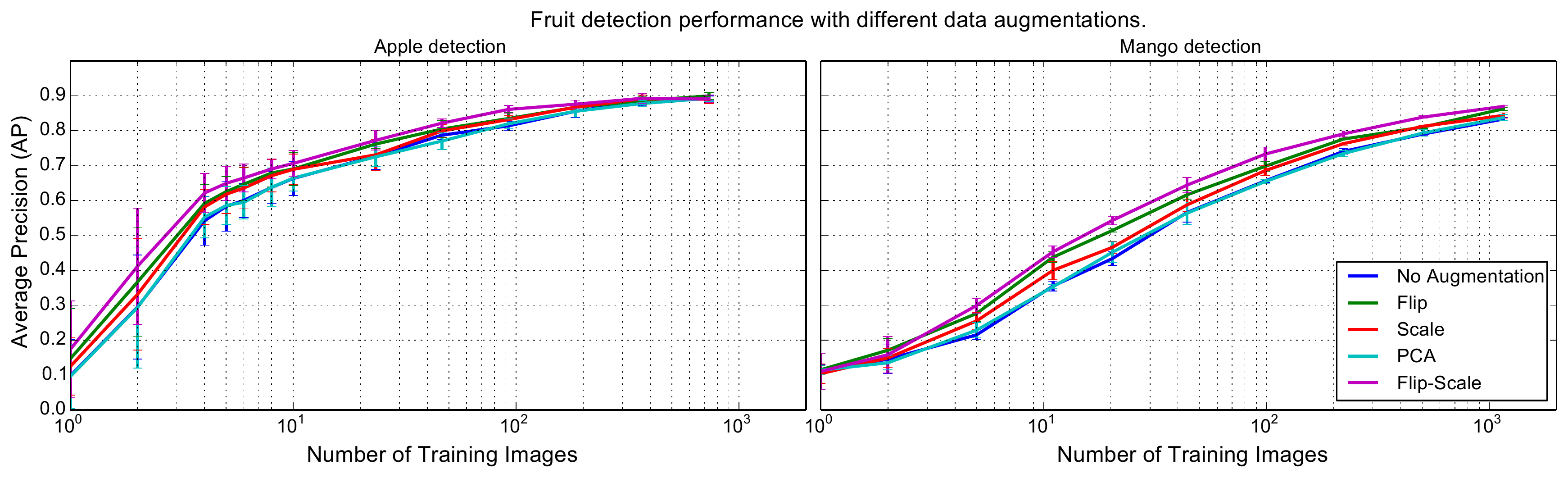}
		\caption{Apple and mango detection performance for different number of training instances with different data augmentation procedures used during training. Best viewed in colour.}
	\label{fig:data-augmentation}
\end{figure*}

Data augmentation was applied with the apple and mango dataset using: flip, scale, flip-scale and PCA augmentations. At each training iteration, scale augmentation rescaled the images to have a shorter size of 300, 500 and 700 pixels, and PCA augmentation perturbed the RGB intensities along each eigenvector by a factor of the eigenvalues multiplied by a uniform random number with zero mean and $0.1$ variance (as done in \cite{Krizhevsky2012}). The results are illustrated in Fig. \ref{fig:data-augmentation}.

PCA augmentation provides negligible improvement (detection results even deteriorated when the augmentation magnitude was increased any further), while both flip and scale augmentations help increase detection performance. The best boost in detection performance is achieved through flip-scale augmentations. This suggests that there is more shape and scale variability in the dataset relative to colour variations along the principal components. Benefits from augmentation decrease as the performance asymptotes with increasing training images. With apples, there is negligible difference at the point of asymptote. whereas for mangoes, the performance asymptote was not reached with the given labelled data (as seen in Fig. \ref{fig:num-training-images}), and so with data augmentation there is still an evident gain in performance. In most cases, with data augmentation, the network reached a fixed detection performance with less than half the number of training samples. For example, with apples, an AP score of $0.86$ is achieved with under $100$ training images, compared to $\sim300$ images required when no augmentation is used. 

\subsection{VGG16 Network}
To obtain peak detection performance for the three fruits, the deeper VGG16 network architecture is trained using all the available training data with flip-scale augmentations. The detection results are reported as fruit F1-scores allowing for comparison against our previous work using the same dataset presented in \cite{Bargoti2016JFR}. The previous approach had used CNNs for pixel-wise classification, followed by watershed segmentation for blob detection. The detection results are shown in Table \ref{tab:fruit-results}.

\begin{table}[h]
  \label{tab:fruit-results}
  \caption{Fruit Detection Results (as F1-scores) with different Faster R-CNN networks and the previously benchmarked pixel-wise CNN architecture.}
  \begin{center}
  \begin{tabular}{cccc} \hline
      Network & Apple & Mango & Almond \Tstrut \\ \hline
      ZF & $0.892$ & $0.876$ & $0.726$ \Tstrut  \\
      VGG16 & $0.904$ & $0.908$ & $0.775$ \\ \hline 
      Pixel-CNN \cite{Bargoti2016JFR} & $0.861$ & $0.836$ & - \Tstrut \\
  \end{tabular}
  \end{center}
\end{table}

The best F1-scores were achieved through the VGG16 Net, with $0.904$, $0.908$ and $0.775$ for the apples, mangoes and almonds respectively. The difference in performance between the ZF and VGG16 network was greatest for mangoes and almonds, where performance had not converged with the number of training images. For both apples and mangoes, Faster R-CNN outperformed the pixel-wise CNN approach. This can be attributed to both the use of a deeper network in Faster R-CNN and the end-to-end approach for detection and localisation (avoiding the heuristic watershed post-processing approach). Fig. \ref{fig:error-detections} shows example detections for each fruit, containing both successful and faulty detections. 

All networks were trained on a Nvidia 980 Ti, using cuda 7.5 and cuDNN 5.1. The VGG16 network took $30-120$ minutes to train for each fruit class, less than the $3$ hours of training time for the pixel-wise CNN based classification system. However, Faster R-CNN performed much faster predictions, with detection on a $500\times500$ image averaging $0.13$ seconds per image with VGG16 net ($0.04$ seconds per image for ZF net) compared to $2-3$ seconds per image for pixel-wise CNN. 

\subsection{Tiled Faster R-CNN}
\begin{figure}[!htb]
	\centering
	\includegraphics[width=0.49\textwidth,clip,trim=0 7 0 10]{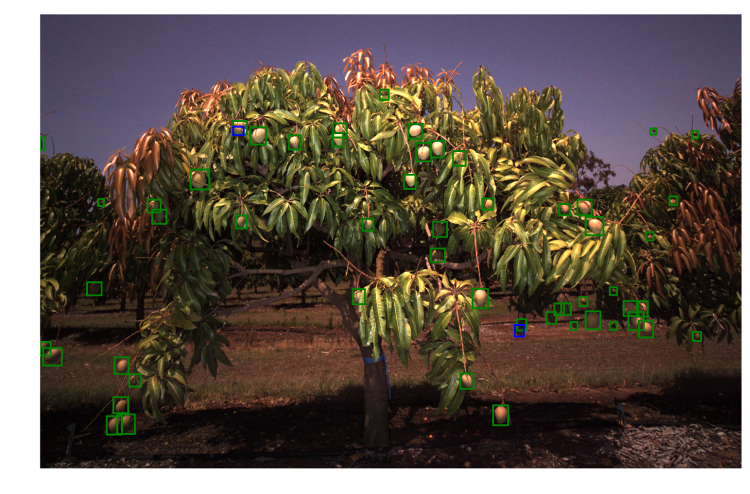}
	\caption{Detection of all mangoes over a mango tree using Tiled Faster R-CNN. The example image contains 54 true positive detections (in green), 2 false negative detections (in blue) and no false positives. Best viewed in colour.}
	\label{fig:whole-tree-detection}
\end{figure}

To perform higher level tasks such as yield mapping and estimation, image prediction needs to be performed over the large (several MP) raw sensor images captured at the farm rather than the sub-images used during training. The GPU memory bottleneck can be overcome by performing detections using smaller sliding windows over the larger images, or `tiling'. Keeping the overlap region greater than the maximum size of the fruit in the data, detections proposals are collected over the sub-sections and thresholding and NMS is applied over fused output on the large image. Fig. \ref{fig:whole-tree-detection} shows fruit detection over a whole tree at the mango orchard, using Tiled Faster R-CNN. The original image ($3496\times2472$) was scanned with smaller windows of $500\times500$ and an overlap of $50$ pixels. The figure shows the image cropped to a single tree that contains $56$ visible fruit. The tiling approach detected $54$ fruit correctly and without any false positive detections. The proposed framework can be used for detecting fruit on trees from images captured across an orchard block. Work conducted in parallel by the authors \cite{Stein2016} use this approach to perform yield estimation and fruit localisation at this mango orchard. 

\section{DISCUSSION}
\label{sec:discussion}

\begin{figure*}[!ht]
	\centering
	\includegraphics[width=1.0\textwidth,clip,trim=0 0 0 0]{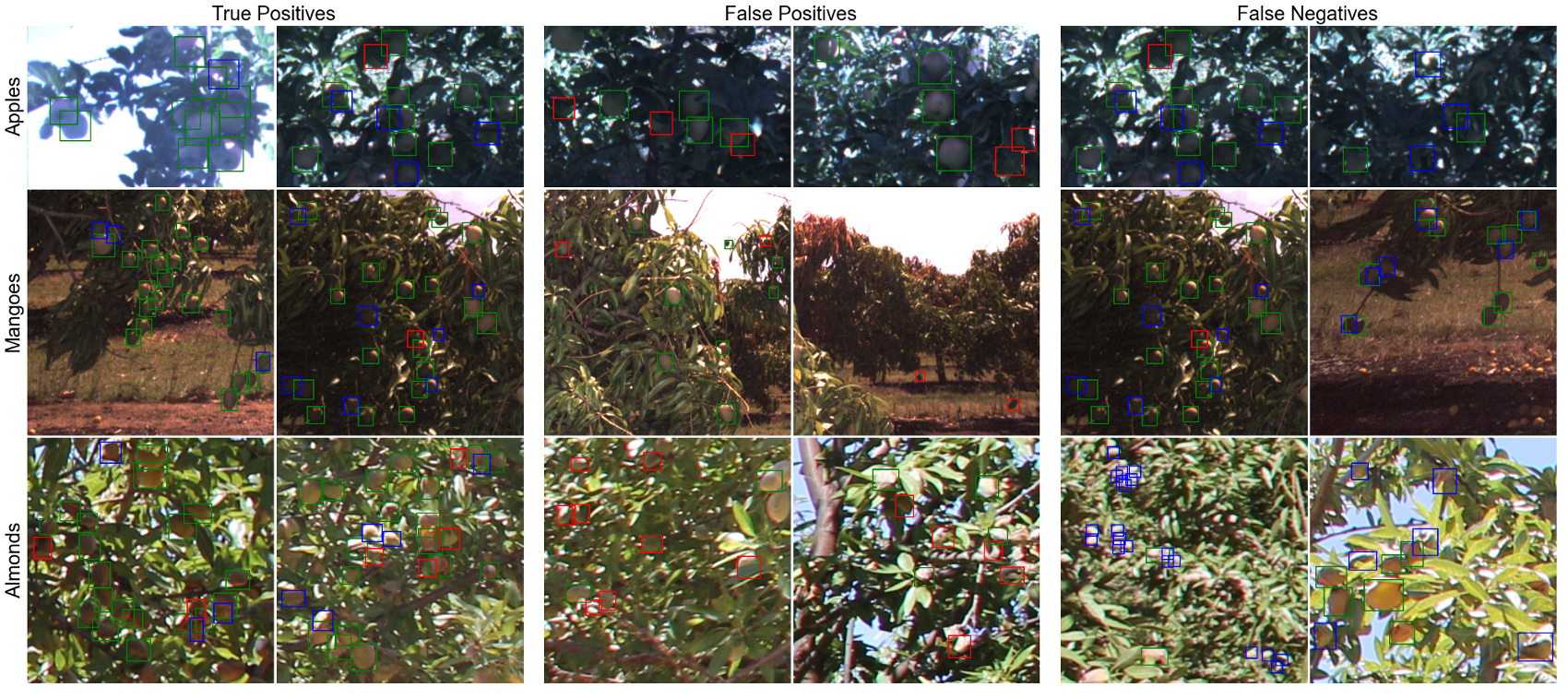}
	\caption{Sample detections over the test set for apples, mangoes and almonds, with true positive detections in green, false positives in red and false negatives in blue. The paired columns contain instances with two examples of the most true positives, most false positives and most false negatives from left to right. Best viewed on screen in colour.}
	\label{fig:error-detections}
\end{figure*}

The Faster R-CNN detection framework yields state-of-the-art performance on multiple orchard image data. This section provides insights into the practical implementation of the detection network in orchards, drawn from the ablations studies presented earlier. 

The study of detection performance against the number training images was useful in identifying which fruit had not reached their performance asymptote and hence may warrant additional data labelling. The $726$ images labelled for apples were enough to span the variability in the dataset. The mango and almond datasets containing $1154$ and $385$ images respectively, however, would require an order of magnitude increase in labelled data to reach that asymptote. Data augmentation was helped boost detection performance, effectively reducing the number of training images required by over $50\%$ and enabling the network to reach the performance asymptote with less labelled data. There was little advantage to transferring knowledge between farms, which is a surprising result given the apparent visual similarities between imagery from different orchards. The ImageNet features were sufficient in performing fine-grained classification and detection in orchards, with the results adding to the increasing evidence of the suitability of ImageNet features for a broad range of image processing tasks \cite{Huh2016}. 

An analysis into fruit precision and recall can give a better insight into the practical significance of the observed improvements in detection performance. Focusing on the mangoes, an improvement in the F1-score from $0.876$ to $0.908$ ($3.2\%$) from the ZF to the VGG16 network equates to a change in precision from $0.933$ to $0.958$ and change in recall from $0.825$ to $0.863$. This means detection of $3.8\%$ extra fruit and a reduction in incorrect detections by $2.5\%$. Depending on the farm application, the network choice is a trade-off between accuracy and speed. For example, yield mapping would benefit from higher detection accuracy and can be performed offline \cite{Bargoti2016JFR}. 

The worst detection performance was observed with the almonds, which can be attributed to aspects of the dataset other than the lower number of training images. The almonds are a smaller fruit on a very large tree, resulting in a low resolution per fruit given the need to capture the whole tree in a single image (Table \ref{tab:dataset-info}). We would require $>30$ MP tree images to get the same pixel density per almond as we have for mangoes and apples. Secondly, almonds were most similar in colour and texture to the foliage, which, when combined with low resolution imagery, resulted in a difficult dataset to manually label and perform detection on. 

The detection approach presented in this paper can be easily extended to other orchards. The provided labelling toolbox can be used for the labelling process, where a parallel training/testing process can help evaluate the change in detection performance with increasing number of training images. The labelling processing could then be terminated based on the labelling budget and/or performance requirements. For smaller fruit, the images need to be rescaled such that the minimum fruit size is greater than $16$ pixels, however, the low fruit resolution can be detrimental for labelling and detection performance. The results advocate the use of simple data augmentation techniques such as image flipping and rescaling, and transfer learning between different fruits is not deemed necessary. Transfer learning could still be important when the base task is very similar to the target task. For example, a model trained at the given apple dataset might still be useful for initialising the detection network for a different apple dataset captured under different illumination conditions and/or with a different sensor. Further tests need to be conducted on how to adapt a model under such variations in the dataset, however, \cite{Sa2016} shows reasonable qualitative performance over a new dataset without any re-learning. 

\subsection{Error Cases}
\label{sec:error-cases}

Fig. \ref{fig:error-detections} shows examples of fruit detections, covering image instances with the most number of true positives, false positives and false negatives. A portion of the detection errors observed can be attributed to, 1) the inability to detect all fruit appearing in a cluster, and 2) error in ground truth labelling resulting in incorrect false positive and false negative evaluations. 

Overlapping detections in clustered regions get suppressed by NMS (seen in image instances in Fig. \ref{fig:error-detections} with a high number of false negatives). To understand the severity of this error, the one-to-one evaluation criteria can be relaxed to allow one detection to represent a cluster. For mango detection with the VGG16 network, this increased the recall from $0.871$ to $0.909$, subsequently increasing the F1-score from $0.907$ to $0.927$. Therefore, $3.8\%$ of the error is from fruits appearing in tight clusters. Although outside the scope of this paper, more recent instance segmentation techniques, which uniquely identify objects in a scene \cite{Ren2016}, may be suitable for fruit disambiguation in clusters. 

With limited image resolution, similarities between fruit and foliage, and inconsistencies in object definition, the labelling task is tedious and prone to errors. 
Missing ground truth annotations are a cause of many of the false positive instances in Fig. \ref{fig:error-detections}. Annotation error can be reduced by consensus voting amongst multiple human labellers, which is an expensive operation. Further investigation is required to test if labelling orchard data is feasible for online job listings such as mechanical turk, as some field expertise is required to discern the fruit from the background. A less expensive means to reduce labelling error would be to use the output from the trained detector to clean-up the ground truth data with a human in the loop, though care would be required to avoid inducing biases.

\section{CONCLUSION}
\label{sec:conclusion}
This paper presented a fruit detection system for image data captured in orchards using the state-of-the-art detection framework, Faster R-CNN. Ablation studies were conducted over three orchard fruit types: apples, mangoes and almonds, to better understand practical deployment of such a system. A study of detection performance against the number of training images demonstrated the amount of training data required to reach convergence. Analysis of transfer learning showed that transferring weights between orchards did not yield significant performance gains over a network initialised directly from the highly generalised ImageNet features. Data augmentation techniques such as flip and scale augmentations were found to improve performance with varying number of training images, resulting in equivalent performance with less than half the number of training images. The study leads to the best yet detection performance in the authors' line of prior work, with an F1-score of $>0.9$ achieved for mangoes and apples. 

For high level applications such as yield mapping and estimation, we proposed Tiled Faster R-CNN to implement a trained model over large images, that are required for fruit counting in orchards. Future work will integrate the detection output from Faster R-CNN with yield mapping, conducting object association between adjacent frames. Additional analysis on fruit detection will also be conducted to understand transfer learning between datasets representing the same fruit, captured over different lighting conditions, sensor configurations and times of the year. 

\section*{ACKNOWLEDGEMENT}
This work is supported by the Australian Centre for Field Robotics at The University of Sydney and by funding from the Australian Government Department of Agriculture and Water Resources as part of its Rural R\&D for profit programme. Thanks to Rishi Ramakrishnan for the insightful discussions on the detection framework and the experimental layout. Further information and videos available at:  \url{http://sydney.edu.au/acfr/agriculture}

\bibliographystyle{IEEEtran}
\bibliography{references}

\end{document}